\documentclass[twocolumn]{cinc}
\usepackage{graphicx}
\usepackage{float}
\usepackage{amsmath}
\usepackage{makecell}

\begin{document}
\bibliographystyle{cinc}

\title{Beat by Beat: \\Classifying Cardiac Arrhythmias with Recurrent Neural Networks}


\author { Patrick Schwab, Gaetano C Scebba, Jia Zhang, Marco Delai, Walter Karlen \\
\ \\ 
 Mobile Health Systems Lab, Department of Health Sciences and Technology \\
 ETH Zurich, Switzerland}

\maketitle

\begin{abstract}
%
%
With tens of thousands of electrocardiogram (ECG) records processed by mobile cardiac event recorders every day, heart rhythm classification algorithms are an important tool for the continuous monitoring of patients at risk. We utilise an annotated dataset of 12,186 single-lead ECG recordings to build a diverse ensemble of recurrent neural networks (RNNs) that is able to distinguish between normal sinus rhythms, atrial fibrillation, other types of arrhythmia and signals that are too noisy to interpret. In order to ease learning over the temporal dimension, we introduce a novel task formulation that harnesses the natural segmentation of ECG signals into heartbeats to drastically reduce the number of time steps per sequence. Additionally, we extend our RNNs with an attention mechanism that enables us to reason about which heartbeats our RNNs focus on to make their decisions. Through the use of attention, our model maintains a high degree of interpretability, while also achieving state-of-the-art classification performance with an average F1 score of 0.79 on an unseen test set (n=3,658).
\end{abstract}

\section{Introduction}

Cardiac arrhythmias are a heterogenous group of conditions that is characterised by heart rhythms that do not follow a normal sinus pattern. One of the most common arrhythmias is atrial fibrillation (AF) with an age-dependant population prevalence of $2.3$ - $3.4\%$ \cite{ball2013atrial}. Due to the increased mortality associated with arrhythmias, receiving a timely diagnosis is of paramount importance for patients \cite{ball2013atrial, camm2010guidelines}. To diagnose cardiac arrhythmias, medical professionals typically consider a patient's electrocardiogram (ECG) as one of the primary factors \cite{camm2010guidelines}. In the past, clinicians recorded these ECGs mainly using multi-lead clinical monitors or Holter devices. However, the recent advent of mobile cardiac event recorders has given patients the ability to remotely record short ECGs using devices with a single lead. 

We propose a machine-learning approach based on recurrent neural networks (RNNs) to differentiate between various types of heart rhythms in this more challenging setting with just a single lead and short ECG record lengths. To ease learning of dependencies over the temporal dimension, we introduce a novel task formulation that harnesses the natural beat-wise segmentation of ECG signals. In addition to utilising several heartbeat features that have been shown to be highly discriminative in previous works, we also use stacked denoising autoencoders (SDAE) \cite{vincent2010stacked} to capture differences in morphological structure. Furthermore, we extend our RNNs with a soft attention mechanism \cite{bahdanau2014neural, xu2015show, yang2016hierarchical,zhang2017mdnet} that enables us to reason about which ECG segments the RNNs prioritise for their decision making.

\section{Methodology}

\begin{figure}[b]
\includegraphics[width=8.2cm]{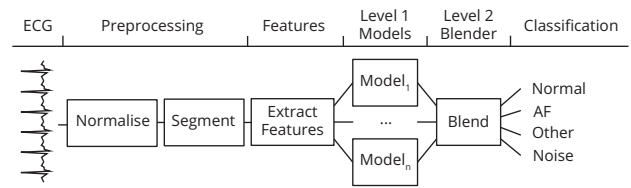}
\caption{An overview of our ECG classification pipeline.}
\label{fig:overview}
\end{figure}

Our cardiac rhythm classification pipeline consists of multiple stages (figure \ref{fig:overview}). The core idea of our setup is to extract a diverse set of features from the sequence of heartbeats in an ECG record to be used as input features to an ensemble of RNNs. We blend the individual models' predictions into a per-class classification score using a multilayer perceptron (MLP) with a softmax output layer. The following paragraphs explain the stages shown in figure \ref{fig:overview} in more detail.

\textbf{ECG Dataset.} We use the dataset of the PhysioNet Computing in Cardiology (CinC) 2017 challenge \cite{Challenge2017} which contains 12,186 unique single-lead ECG records of varying length. Experts annotated each of these ECGs as being either a normal sinus rhythm, AF, an other arrhythmia or too noisy to classify. The challenge organisers keep 3,658 ($30\%$) of these ECG records private as a test set. Additionally, we hold out a non-stratified random subset of $20\%$ of the public dataset as a validation set. For some RNN configurations, we further augment the training data with labelled samples extracted from other PhysioNet databases \cite{Goldbergere215, moody2001impact, moody1983new, greenwald1990improved} in order to even out misbalanced class sizes in the training set. As an additional measure against the imbalanced class distribution of the dataset, we weight each training sample's contribution to the loss function to be inversely proportional to its class' prevalence in the overall dataset.

\textbf{Normalisation.} Prior to segmentation, we normalise the ECG recording to have a mean value of zero and a standard deviation of one. We do not apply any additional filters as all ECGs were bandpass-filtered by the recording device. 

\textbf{Segmentation.} Following normalisation, we segment the ECG into a sequence of heartbeats. We decide to reformulate the given task of classifying arrhythmias as a sequence classification task over heartbeats rather than over raw ECG readings. The motivation behind the reformulation is that it significantly reduces the number of time steps through which the error signal of our RNNs has to propagate. On the training set, the reformulation reduces the mean number of time steps per ECG from $9000$ to just $33$. To perform the segmentation, we use a customised QRS detector based on Pan-Tompkin's \cite{pan1985real} that identifies R-peaks in the ECG recording. We extend their algorithm by adapting the threshold with a moving average of the ECG signal to be more resilient against the commonly encountered short bursts of noise. For the purpose of this work, we define heartbeats using a symmetric fixed size window with a total length of $0.66$ seconds around R-peaks. We pass the extracted heartbeat sequence in its original order to the feature extraction stage.

\textbf{Feature Extraction.} We extract a diverse set of features from each heartbeat in an ECG recording. Specifically, we extract the time since the last heartbeat ($\delta$RR), the relative wavelet energy (RWE) over five frequency bands, the total wavelet energy (TWE) over those frequency bands, the R amplitude, the Q amplitude, QRS duration and wavelet entropy (WE). Previous works demonstrated the efficacy of all of these features in discriminating cardiac arrhythmias from normal heart rhythms \cite{sarkar2008detector, tateno2001automatic, garcia2016application, rodenas2015wavelet, alcaraz2006wavelet}. In addition to the aforementioned features, we also train two SDAEs on the heartbeats in an unsupervised manner with the goal of learning more nuanced differences in morphology of individual heartbeats. We train one SDAE on the extracted heartbeats of the training set and the other on their  wavelet coefficients. We then use the encoding side of the SDAEs to extract low-dimensional embeddings of each heartbeat and each heartbeat's wavelet coefficients to be used as additional input features. Finally, we concatenate all extracted features into a single feature vector per heartbeat and pass them to the level 1 models in original heartbeat sequence order.

\textbf{Level 1 Models.} We build an ensemble of level 1 models to classify the sequence of per-beat feature vectors. To increase the diversity within our ensemble, we train RNNs in various binary classification settings and with different hyperparameters. We use RNNs with $1$ - $5$ recurrent layers that consist of either Gated Recurrent Units (GRU) \cite{chung2014empirical} or Bidirectional Long Short-Term Memory (BLSTM) units \cite{graves2013hybrid}, followed by an optional attention layer, $1$ - $2$ forward layers and a softmax output layer. Additionally, we infer a nonparametric Hidden Semi-Markov Model (HSMM) \cite{johnson2013bayesian} with $64$ initial states for each class in an unsupervised setting. In total, our ensemble of level 1 models consists of 15 RNNs and 4 HSMMs. We concatenate the ECG's normalised log-likelihoods under the per-class HSMMs and the RNNs' softmax outputs into a single prediction vector. We pass the prediction vector of the level 1 models to the level 2 blender model.

\textbf{Level 2 Blender.} We use blending \cite{wolpert1992stacked} to combine the predictions of our level 1 models and a set of ECG-wide features into a final per-class classification score. The additional features are the RWE and WE over the whole ECG and the absolute average deviation (AAD) of the WE and $\delta$RR of all beats. We employ a MLP with a softmax output layer as our level 2 blender model. In order to avoid overfitting to the training set, we train the MLP on the validation set.

\textbf{Hyperparameter Selection.} To select the hyperparameters of our level 1 RNNs, we performed a grid search on the range of $0$ - $75\%$ for the dropout and recurrent dropout percentages, $60$ - $512$ for the number of units per hidden layer and $1$ - $5$ for the number of recurrent layers. We found that RNNs trained with $35\%$ dropout, $65\%$ recurrent dropout, $80$ units per hidden layer and $5$ recurrent layers (plus an additional attention layer) achieve consistently strong results across multiple binary classification settings. For our level 2 blender model, we utilise Bayesian optimisation \cite{bergstra2013hyperopt} to select the number of layers, number of hidden units per layer, dropout and number of training epochs. We perform a 5-fold cross validation on the validation set to select the blender model's hyperparameters.

\subsection{Attention over Heartbeats}
\begin{figure*}[t]
\centering
\includegraphics[trim={-0.23cm 0cm -0.55cm 0cm},clip,width=7.85cm]{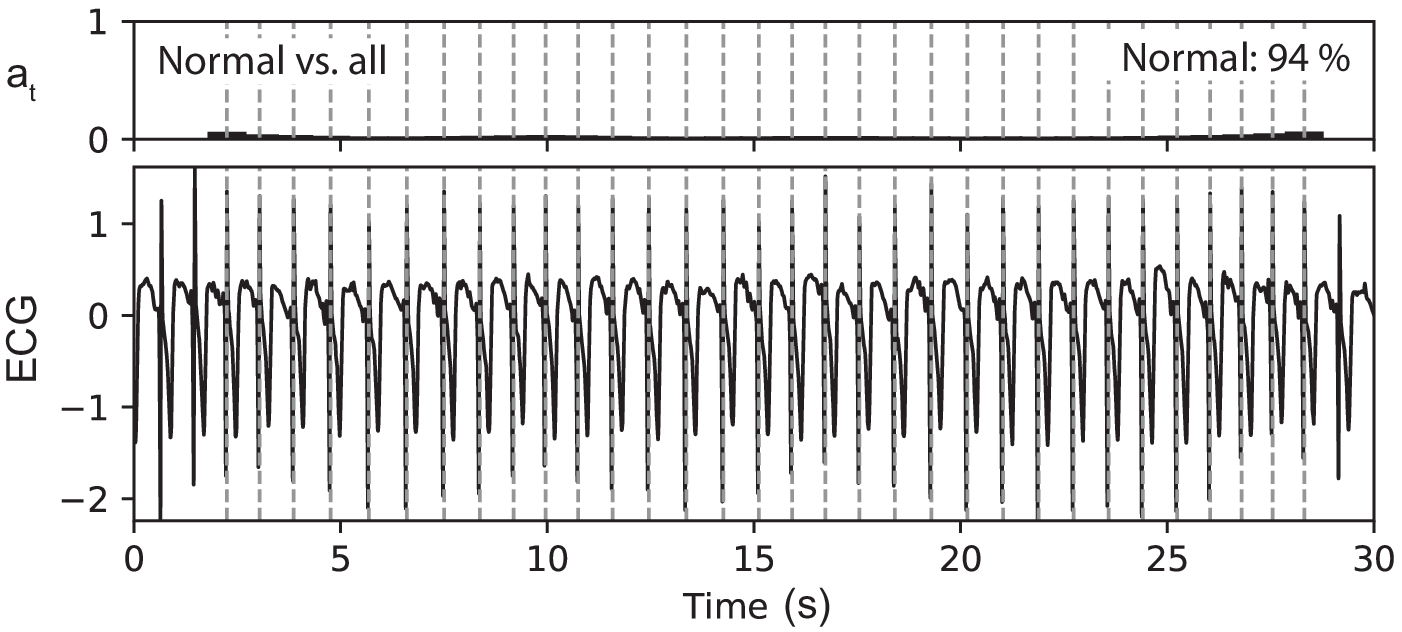}
\includegraphics[trim={1cm 0cm 1cm 5.5cm},clip,width=7.4cm]{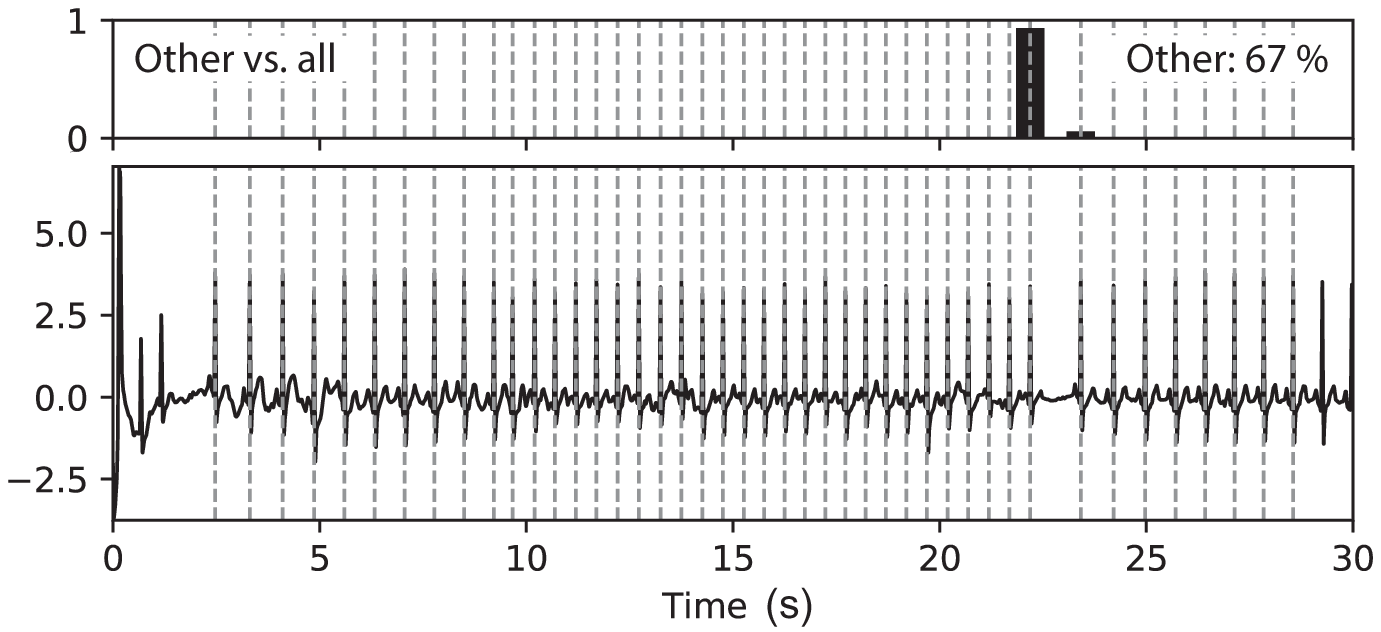}
\caption{A visualisation of the attention values $a_t$ (top) of two different RNNs over two sample ECG recordings (bottom). The graphs on top of the ECG recordings show the attention values $a_{t}$ associated with each identified heartbeat (dashed line). The labels in the left and right corners of the attention value graphs show the settings the model was trained for and their classification confidence, respectively. The recording on the left (A02149) represents a normal sinus rhythm. Due to the regular heart rhythm in the ECG, a distinctive pattern of approximately equally weighted attention on each heartbeat emerges from our RNN that was trained to distinguish between normal sinus rhythms and all other types of rhythms. The recording on the right (A04661) is labelled as an other arrhythmia. The RNN trained to identify other arrhythmias focuses primarily on a sudden, elongated pause in the heart rhythm to decide that the record is most likely an other arrhythmia.}
\label{fig:attention}
\end{figure*}

Attention mechanisms have been shown to allow for greater interpretability of neural networks in a variety of tasks in computer vision and natural language processing \cite{bahdanau2014neural, xu2015show, yang2016hierarchical, zhang2017mdnet}. In this work, we apply soft attention over the heartbeats contained in ECG signals in order to gain a deeper understanding of the decision-making process of our RNNs. Consider the case of an RNN that is processing a sequence of $T$ heartbeats. The topmost recurrent layer outputs a hidden state $h_{t}$ at every time step $t \in [1, T]$ of the sequence. We extend some of our RNNs with additive soft attention over the hidden states $h_{t}$ to obtain a context vector $c$ that attenuates the most informative hidden states $h_{t}$ of a heartbeat sequence. Based on the definition in \cite{yang2016hierarchical}, we use the following set of equations:
\begin{align}
u_{t} &= tanh(W_{beat}h_{t} + b_{beat}) \\
a_{t} &= softmax(u_{t}^Tu_{beat}) \\
c &= \sum_{t}a_{t}h_{t} 
\end{align}
Where equation (1) is a single-layer MLP with a weight matrix $W_{beat}$ and bias $b_{beat}$ to obtain $u_t$ as a hidden representation of $h_t$ \cite{yang2016hierarchical}. In equation (2), we calculate the attention factors $a_t$ for each heartbeat by computing a softmax over the dot-product similarities of every heartbeat's $u_t$ to the heartbeat context vector $u_{beat}$. $u_{beat}$ corresponds to a hidden representation of the most informative heartbeat \cite{yang2016hierarchical}. We jointly optimise $W_{beat}$, $b_{beat}$ and $u_{beat}$ with the other RNN parameters during training. In figure \ref{fig:attention}, we showcase two examples of how qualitative analysis of the attention factors $a_t$ of equation (2) provides a deeper understanding of our RNNs' decision making.

\section{Related Work}

Our work builds on a long history of research in detecting cardiac arrhythmias from ECG records by making use of features that have been shown to be highly discriminative in distinguishing certain arrhythmias from normal heart rhythms \cite{sarkar2008detector, tateno2001automatic, garcia2016application, rodenas2015wavelet, alcaraz2006wavelet}. Recently, Rajpurkar et al. proposed a 34-layer convolutional neural network (CNN) to reach cardiologist-level performance in classifying a large set of arrhythmias from mobile cardiac event recorder data \cite{rajpurkar2017cardiologist}. In contrast, we achieve state-of-the-art performance with significantly fewer trainable parameters by harnessing the natural heartbeat segmentation of ECGs and discriminative features from previous works. Additionally, we pay consideration to the fact that interpretability remains a challenge in applying machine learning to the medical domain \cite{Unintended2017} by extending our models with an attention mechanism that enables medical professionals to reason about which heartbeats contributed most to the decision-making process of our RNNs.

\section{Results and Conclusion}

We present a machine-learning approach to distinguishing between multiple types of heart rhythms. Our approach utilises an ensemble of RNNs to jointly identify temporal and morphological patterns in segmented ECG recordings of any length. In detail, our approach reaches an average F1 score of 0.79 on the private test set of the PhysioNet CinC Challenge 2017 ($n=3,658$) with class-wise F1 scores of $0.90$, $0.79$ and $0.68$ for normal rhythms, AF and other arrhythmias, respectively. On top of its state-of-the-art performance, our approach maintains a high degree of interpretability through the use of a soft attention mechanism over heartbeats. In the spirit of open research, we make an implementation of our cardiac rhythm classification system available through the PhysioNet 2017 Open Source Challenge.

\textbf{Future Work.} Based on our discussions with a cardiologist, we hypothesise that the accuracy of our models could be further improved by incorporating contextual information, such as demographic information, data from other clinical assessments and behavioral aspects. 

\section*{Acknowledgements}
%
This work was partially funded by the Swiss National Science Foundation (SNSF) project No. 167302 within the National Research Program (NRP) $75$ ``Big Data'' and SNSF project No. 150640. We thank Prof. Dr. med. Firat Duru for providing valuable insights into the decision-making process of cardiologists.

\balance
\bibliography{refs}

\begin{thebibliography}{10}
\expandafter\ifx\csname url\endcsname\relax
  \def\url#1{\texttt{#1}}\fi
\expandafter\ifx\csname urlprefix\endcsname\relax\def\urlprefix{URL }\fi

\bibitem{ball2013atrial}
Ball J, Carrington MJ, McMurray JJ, Stewart S.
\newblock Atrial fibrillation: Profile and burden of an evolving epidemic in
  the 21st century.
\newblock International Journal of Cardiology
  2013;\hspace{0pt}167(5):1807--1824.

\bibitem{camm2010guidelines}
Camm AJ, Kirchhof P, Lip GY, Schotten U, Savelieva I, Ernst S, Van~Gelder IC,
  Al-Attar N, Hindricks G, Prendergast B, et~al.
\newblock Guidelines for the management of atrial fibrillation.
\newblock European Heart Journal 2010;\hspace{0pt}31:2369--–2429.

\bibitem{vincent2010stacked}
Vincent P, Larochelle H, Lajoie I, Bengio Y, Manzagol PA.
\newblock Stacked denoising autoencoders: Learning useful representations in a
  deep network with a local denoising criterion.
\newblock Journal of Machine Learning Research
  2010;\hspace{0pt}11(Dec):3371--3408.

\bibitem{bahdanau2014neural}
Bahdanau D, Cho K, Bengio Y.
\newblock Neural machine translation by jointly learning to align and
  translate.
\newblock In International Conference on Learning Representations, 2015.

\bibitem{xu2015show}
Xu K, Ba J, Kiros R, Cho K, Courville A, Salakhudinov R, Zemel R, Bengio Y.
\newblock Show, attend and tell: Neural image caption generation with visual
  attention.
\newblock In International Conference on Machine Learning. 2015;\hspace{0pt}
  2048--2057.

\bibitem{yang2016hierarchical}
Yang Z, Yang D, Dyer C, He X, Smola AJ, Hovy EH.
\newblock Hierarchical attention networks for document classification.
\newblock In Conference of the North American Chapter of the Association for
  Computational Linguistics: Human Language Technologies. 2016;\hspace{0pt}
  1480--1489.

\bibitem{zhang2017mdnet}
Zhang Z, Xie Y, Xing F, McGough M, Yang L.
\newblock {MDNet: A Semantically and Visually Interpretable Medical Image
  Diagnosis Network}.
\newblock {In International Conference on Computer Vision and Pattern
  Recognition, arXiv preprint arXiv:1707.02485}, 2017.

\bibitem{Challenge2017}
Clifford GD, Liu CY, Moody B, Lehman L, Silva I, Li Q, Johnson AEW, Mark RG.
\newblock {AF classification from a short single lead ECG recording: The
  Physionet Computing in Cardiology Challenge 2017}.
\newblock In Computing in Cardiology, 2017.

\bibitem{Goldbergere215}
Goldberger AL, Amaral LAN, Glass L, Hausdorff JM, Ivanov PC, Mark RG, Mietus
  JE, Moody GB, Peng CK, Stanley HE.
\newblock {PhysioBank, PhysioToolkit, and PhysioNet: Components of a New
  Research Resource for Complex Physiologic Signals}.
\newblock Circulation 2000;\hspace{0pt}101(23):e215--e220.

\bibitem{moody2001impact}
Moody GB, Mark RG.
\newblock {The impact of the MIT-BIH arrhythmia database}.
\newblock IEEE Engineering in Medicine and Biology Magazine
  2001;\hspace{0pt}20(3):45--50.

\bibitem{moody1983new}
Moody G.
\newblock {A new method for detecting atrial fibrillation using RR intervals}.
\newblock In Computers in Cardiology. IEEE, 1983;\hspace{0pt} 227--230.

\bibitem{greenwald1990improved}
Greenwald SD, Patil RS, Mark RG.
\newblock Improved detection and classification of arrhythmias in
  noise-corrupted electrocardiograms using contextual information.
\newblock In Computers in Cardiology. IEEE, 1990;\hspace{0pt} 461--464.

\bibitem{pan1985real}
Pan J, Tompkins WJ.
\newblock A real-time {QRS} detection algorithm.
\newblock IEEE Transactions on Biomedical Engineering
  1985;\hspace{0pt}3:230--236.

\bibitem{sarkar2008detector}
Sarkar S, Ritscher D, Mehra R.
\newblock A detector for a chronic implantable atrial tachyarrhythmia monitor.
\newblock IEEE Transactions on Biomedical Engineering
  2008;\hspace{0pt}55(3):1219--1224.

\bibitem{tateno2001automatic}
Tateno K, Glass L.
\newblock Automatic detection of atrial fibrillation using the coefficient of
  variation and density histograms of {RR} and $\delta${RR} intervals.
\newblock Medical and Biological Engineering and Computing
  2001;\hspace{0pt}39(6):664--671.

\bibitem{garcia2016application}
Garc{\'\i}a M, R{\'o}denas J, Alcaraz R, Rieta JJ.
\newblock Application of the relative wavelet energy to heart rate independent
  detection of atrial fibrillation.
\newblock computer methods and programs in biomedicine
  2016;\hspace{0pt}131:157--168.

\bibitem{rodenas2015wavelet}
R{\'o}denas J, Garc{\'\i}a M, Alcaraz R, Rieta JJ.
\newblock Wavelet entropy automatically detects episodes of atrial fibrillation
  from single-lead electrocardiograms.
\newblock Entropy 2015;\hspace{0pt}17(9):6179--6199.

\bibitem{alcaraz2006wavelet}
Alcaraz R, Vay{\'a} C, Cervig{\'o}n R, S{\'a}nchez C, Rieta J.
\newblock Wavelet sample entropy: A new approach to predict termination of
  atrial fibrillation.
\newblock In Computers in Cardiology. IEEE, 2006;\hspace{0pt} 597--600.

\bibitem{chung2014empirical}
Chung J, Gulcehre C, Cho K, Bengio Y.
\newblock Empirical evaluation of gated recurrent neural networks on sequence
  modeling.
\newblock In Neural Information Processing Systems, Workshop on Deep Learning,
  arXiv preprint arXiv:1412.3555, 2014.

\bibitem{graves2013hybrid}
Graves A, Jaitly N, Mohamed Ar.
\newblock Hybrid speech recognition with deep bidirectional lstm.
\newblock In {Automatic Speech Recognition and Understanding, IEEE Workshop
  on}. IEEE, 2013;\hspace{0pt} 273--278.

\bibitem{johnson2013bayesian}
Johnson MJ, Willsky AS.
\newblock Bayesian nonparametric hidden semi-markov models.
\newblock Journal of Machine Learning Research
  2013;\hspace{0pt}14(Feb):673--701.

\bibitem{wolpert1992stacked}
Wolpert DH.
\newblock Stacked generalization.
\newblock Neural networks 1992;\hspace{0pt}5(2):241--259.

\bibitem{bergstra2013hyperopt}
Bergstra J, Yamins D, Cox DD.
\newblock {Hyperopt: A python library for optimizing the hyperparameters of
  machine learning algorithms}.
\newblock In Proceedings of the 12th Python in Science Conference.
  2013;\hspace{0pt} 13--20.

\bibitem{rajpurkar2017cardiologist}
Rajpurkar P, Hannun AY, Haghpanahi M, Bourn C, Ng AY.
\newblock Cardiologist-level arrhythmia detection with convolutional neural
  networks.
\newblock {arXiv preprint, arXiv:1707.01836}, 2017.

\bibitem{Unintended2017}
Cabitza F, Rasoini R, Gensini G.
\newblock Unintended consequences of machine learning in medicine.
\newblock Journal of the American Medical Association
  2017;\hspace{0pt}318(6):517--518.

\end{thebibliography}
\begin{correspondence}
Patrick Schwab, ETH Zurich\\
Balgrist Campus, BAA D, Lengghalde 5, 8092 Zurich\\
patrick.schwab@hest.ethz.ch
\end{correspondence}

\end{document}